\documentclass{llncs}
\usepackage{llncsdoc}
\usepackage{amsmath, amsfonts}
\usepackage[utf8]{inputenc}

\newtheorem{hypothesis}{Hypothesis}

\newcommand{\bpf} {\noindent{\sc Proof} : }
\newcommand{\epf} {\hfill$\square$\vspace{.5cm}}
\newcommand{\E} {\mathbb{E}}

\renewcommand{\P} {\mathbb{P}}

\newcommand{\N}{\mathbb{N}}
\newcommand{\R}{\mathbb{R}}

\newcommand{\h}{\mathcal{H}}

\newcommand {\ep} {\varepsilon}
\newcommand {\x} {\mathcal{X}}
\newcommand {\y} {\mathcal{Y}}
\newcommand {\I} {\mathcal{I}}
\newcommand {\zi} {{Z^i}}
\renewcommand {\l} {\ell}

\DeclareMathOperator*{\argmin}{arg\,min}

\usepackage{color}
\definecolor{MyDarkBlue}{rgb}{0.05,0.,0.8}

\title{
Stability of Multi-Task Kernel Regression Algorithms}

\titlerunning{Stability of Multi-Task Kernel Regression}

\author{Julien Audiffren  \and Hachem Kadri}

\authorrunning{J. Audiffren and H. Kadri }

\institute{QARMA, LIF/CNRS, Aix-Marseille University}

\begin{document}

\maketitle

\begin{abstract}

{
{We study the stability properties of nonlinear multi-task regression in reproducing Hilbert spaces with operator-valued kernels. 
Such kernels, a.k.a. multi-task kernels, are appropriate for learning problems with nonscalar outputs like multi-task learning and structured output prediction. We show that multi-task kernel regression algorithms are uniformly stable in the general case of infinite-dimensional output spaces.  We then derive under mild assumption on the kernel generalization bounds of such algorithms, and we show their consistency even with non Hilbert-Schmidt operator-valued kernels\footnote{See Definition~\ref{noyau HS} for a precise statement of what we mean by Hilbert-Schmidt operator-valued kernel.}. We demonstrate how to apply the results to various multi-task kernel regression methods such as vector-valued SVR and functional ridge regression.}

}
\end{abstract}

\section{Introduction}

{A central issue in the field of machine learning is to design and analyze the generalization ability of learning algorithms. Since the seminal work of Vapnik and Chervonenkis~\cite{vapnik71}, various approaches and techniques have been advocated and a large body of literature has emerged in learning theory providing rigorous generalization and performance bounds~\cite{Herbrich02}. This literature has mainly focused on scalar-valued function learning algorithms like binary classification~\cite{Boucheron05} and real-valued regression~\cite{Gyorfi02}. 
However, interest in learning vector-valued functions is increasing~\cite{micchelli05}. Much of this interest stems from the need for more sophisticated learning methods suitable for complex-output learning problems such as multi-task learning~\cite{caruana97} and structured output prediction~\cite{bakir07}. 
Developing generalization bounds for vector-valued function learning algorithms then becomes more and more crucial to the theoretical understanding of such complex algorithms.
Although relatively recent, the effort in this area has already produced several successful results, 
including~\cite{Baxter00,Maurer05,Maurer06,Ando05}. 
Yet, these studies have considered only the case of finite-dimensional output spaces, and have focused more on linear machines than nonlinear ones. 
To the best of our knowledge, the only work investigating the generalization performance of nonlinear multi-task learning methods when output spaces can be infinite-dimensional is that of Caponnetto and De Vito~\cite{VitoCapo2006}. In their study, the authors have derived from a theoretical (minimax) analysis generalization bounds for regularized least squares regression in reproducing kernel Hilbert spaces~(RKHS) with operator-valued kernels. {It should be noted that}, unlike the scalar-valued function learning setting, the reproducing kernel in this context is a positive-definite operator-valued function\footnote{The kernel is a matrix-valued function in the case of finite dimensional output spaces.}. The operator has the advantage of allowing us to take into account dependencies between different tasks and then to model task relatedness. Hence, these kernels are known to extend linear multi-task learning methods to the nonlinear case, and are referred to as \textit{multi-task kernels}\footnote{In the context of this paper, operator-valued kernels and multi-task kernels mean the same thing.}~\cite{Micchelli-2005b,Evgeniou05}. 

The convergence rates proposed by Caponnetto and De Vito~\cite{VitoCapo2006}, although optimal in the case of finite-dimensional output spaces, require assumptions on the kernel that can be restrictive in the infinite-dimensional case. 
Indeed, their proof depends upon the fact that the kernel is Hilbert-Schmidt~(see Definition~\ref{noyau HS}) and this restricts the applicability of their results when the output space is infinite-dimensional. To illustrate this, let us consider the identity operator-based multi-task kernel $K(\cdot,\cdot) = k(\cdot,\cdot) I$, where $k$ is a scalar-valued kernel and $I$ is the identity operator. 
This kernel which was already used by Brouard et al.~\cite{brouard11} and Grunewalder et al.~\cite{arthur12} for structured output prediction and conditional mean embedding, respectively, does not satisfy the Hilbert-Schmidt assumption~(see Remark~\ref{remarkHS}), and therefore the results of~\cite{VitoCapo2006} cannot be applied in this case~(for more details see Section~\ref{sectionExample}). 
It is also important to note that, since the analysis of Caponnetto and De Vito~\cite{VitoCapo2006} is based on a measure of the complexity of the hypothesis space independently of the algorithm, it does not take into account the properties of learning algorithms. 

In this paper, we address these issues by studying the stability of multi-task kernel regression algorithms when the output space is a~(possibly infinite-dimensional) Hilbert space. The notion of \textit{algorithmic stability}, which is the behavior of a learning algorithm following a change of the training data, was used successfully by Bousquet and Elisseeff~\cite{Bousquet} to derive bounds on the generalization error of deterministic scalar-valued learning algorithms.
Subsequent studies extended this result to cover other learning algorithms such as randomized, transductive and ranking algorithms~\cite{Elisseeff05,Cortes08,agarwal09}, both in i.i.d\footnote{The abbreviation ``i.i.d.'' stands for ``independently and identically distributed'' } and non-i.i.d scenarios~\cite{Mohri10}. But, none of these papers is directly concerned with  the stability of nonscalar-valued learning algorithms. 
%
It is the aim of the present work to extend the stability results of~\cite{Bousquet} to cover vector-valued learning schemes associated with multi-task kernels.
Specifically, we make the following contributions in this paper: \textbf{1)} we show that multi-task kernel regression algorithms are uniformly stable for the general case of infinite-dimensional output spaces,  \textbf{2)} we derive under mild assumption on the kernel generalization bounds of such algorithms, and we show their consistency even with non Hilbert-Schmidt operator-valued kernels (see Definition~\ref{noyau HS}), \textbf{3)} we demonstrate how to apply these results to various multi-task regression methods such as vector-valued support vector regression (SVR) and functional ridge regression,  \textbf{4)} we provide examples of infinite-dimensional multi-task kernels which are not Hilbert-Schmidt, showing that our assumption on the kernel is weaker than the one in \cite{VitoCapo2006}. 

The rest of this paper is organized as follows. In Section~\ref{sectionHypothesis} we introduce the necessary notations and briefly recall the main concepts of operator-valued kernels and the corresponding Hilbert-valued RKHS. Moreover, we describe in this section the mathematical assumptions required by the subsequent developments. In Section~\ref{sectionStability} we state the result establishing the stability and providing the generalization bounds of multi-task kernel based learning algorithms. {In Section~\ref{section::StableAlgo}, we show that many
existing multi-task kernel regression algorithms such as vector-valued SVR and functional ridge regression do satisfy the stability requirements. In Section~\ref{sectionExample} we give examples of non Hilbert-Schmidt operator-valued kernels that illustrate the usefulness of our result. 
Section~\ref{sectionConclusion} concludes the paper.

}

\section{Notations, Background and Assumptions}\label{sectionHypothesis}
In this section we introduce the notations we will use in this paper. Let $(\Omega, \mathcal{F},\P)$ be a probability space, $\x$ a Polish space, $\y$ a (possibly infinite-dimensional) separable Hilbert space, $\h$ a separable Reproducing Kernel Hilbert Space (RKHS)~$\subset \y^\x$ with $K$ its reproducing kernel, and $L(\y)$\footnote{We denote by $My=M(y)$ the application of the operator $M \in L(\y)$ to $y\in\y$.}  the space of continuous endomorphisms of $\y$ equipped with the operator norm. Let $\lambda >0$ and $(X_1,Y_1),....,(X_m,Y_m)$ be $m$ i.i.d. copies of the pair of random variables $(X,Y)$ following the unknown distribution~$P$.

We consider a training set $Z=\left\{(x_1,y_1),....,(x_m,y_m) \right\}$ consisting of a realization of~$m$ i.i.d. copies of $(X,Y)$, and we denote by $\zi =Z \setminus (x_i,y_i)$ the set $Z$ where the couple $(x_i,y_i)$ is removed.
Let $c:\y \times \h \times \x \rightarrow \R^+$ be a loss function. We will describe stability and consistency results in Section~\ref{sectionStability}  for a general loss function, while  in Section~\ref{sectionExample} we will provide examples to illustrate them with specific forms of $c$.
The goal of multi-task kernel regression is to find a function $f$, $\h \ni f: \x \rightarrow \y$,  that minimizes a risk functional $$R(f) = \int c(y,f(x),x)  dP(x,y).$$
The empirical risk of $f$ on $Z$ is then
\begin{equation}\nonumber
R_{emp}(f,Z)=\frac{1}{m}\sum_{k=1}^m c(y_k,f,x_k) ,
\end{equation}
and its regularized version is given by
\begin{equation}\nonumber
R_{reg}(f,Z) = R_{emp}(f,Z) +  \lambda \| f \|_\h^2.
\end{equation}
We will denote by 
\begin{equation}\label{definition fZ}
f_Z = \argmin_{f \in \h} R_{reg}(f,Z),
\end{equation}
the function minimizing the regularized risk over $\h$.

Let us now recall the definition of the operator-valued kernel $K$ associated to the RKHS $\h$ when $\y$ is infinite dimensional. For more details see~\cite{micchelli05}.
\begin{definition}\label{noyau HS}
The application $K : \x \times \x \rightarrow L(\y) $ is called the Hermitian positive-definite reproducing operator-valued kernel of the RKHS $\h$ if and only if :
\begin{enumerate}
\item $\forall x \in \x,\forall y \in \y,$ the application 
\begin{equation}\nonumber
\begin{aligned}
 K(.,x)y : \x &\to \y \\
 			x' &\mapsto K(x',x)y
\end{aligned}
\end{equation} 
belongs to $\h$.

\item $\forall f \in \h$, $\forall x \in \x,$ $\forall y \in \y,$
$$ \left\langle f(x),y \right \rangle_\y = \left\langle f,K(.,x)y \right \rangle_\h, $$
\item $\forall x_1,x_2 \in \x,$ $$K(x_1,x_2)=K(x_2,x_1)^* \in L(\y),  \text{ ($*$ denotes the adjoint)}$$
\item $\forall n \ge 1,$ $\forall (x_i,i\in \{1..n\}), (x'_i,i\in \{1..n\}) \in \x^n$, $\forall (y_i,i\in \{1..n\}), (y'_i,i\in \{1..n\}) \in \y^n$,
$$ \sum_{k,\l=0}^n \left\langle K(.,x_k)y_k,K(.,x'_\l)y'_\l \right \rangle_\y  \ge 0.$$
\end{enumerate}
(i) and (ii) define a reproducing kernel, (iii) and (iv) corresponds to the Hermitian and  positive-definiteness properties, respectively.
\\[0.2cm]
Moreover, the kernel $K$ will be called Hilbert-Schmidt if and only if $\forall x \in \x$, $\exists (y_i)_{i\in \N}$, a base of $\y$, such that $Tr(K(x,x))= \sum_{i \in \N} \vert \left\langle K(x,x)y_i,y_i \right\rangle_\h \vert < \infty. $ This is equivalent to saying that the operator $K(x,x) \in L(\y)$ is Hilbert-Schmidt.
\end{definition}

We now discuss the main assumptions we need to prove our results. We start by the following hypothesis on the kernel $K$.
\begin{hypothesis}\label{Hyp K bounded}
$\exists \kappa >0$ such that $\forall x \in \x$, $$\| K(x,x) \|_{op} \le \kappa ^ 2,$$
where $\displaystyle\| K(x,x) \|_{op}= \sup_{y \in \y} \frac{\|K(x,x)y\|_\y} {\|y\|_\y}$ is the operator norm of $K(x,x)$ on $L(\y)$.
\end{hypothesis}

\begin{remark}
\label{remarkHS}

 It is important to note that Hypothesis~\ref{Hyp K bounded} is weaker than the one used in~\cite{VitoCapo2006} which requires that $K$ is Hilbert-Schmidt and $\sup_{x \in \x} Tr(K(x,x)) < + \infty $.
 While the two assumptions are equivalent when the output space $\y$ is finite-dimensional, this is no longer the case when, as in this paper,  $\dim\y=+\infty$.
 Moreover, we can observe that if the hypothesis of~\cite{VitoCapo2006} is satisfied,  then our Hypothesis~\ref{Hyp K bounded} holds (see proof below). The converse is not true (see Section~\ref{sectionExample} for some counterexamples).
\end{remark}

{\noindent{\sc Proof of Remark 1} : }
Let $K$ be a multi-task kernel satisfying the hypotheses of \cite{VitoCapo2006}, i.e $K$ is Hilbert-Schmidt and $\sup_{x \in \x} Tr(K(x,x)) < + \infty $. 
Then, $\exists \eta >0$, $\forall x\in \x$, $\exists \left(e_j^x\right)_{j\in\N}$ an orthonormal basis of $\y$, $\exists \left(h_j^x\right)_{j\in\N}$ an orthogonal family of $\h$ with $\sum_{j\in\N} \|h_j^x\|_\h^2 \le \eta$ such that $\forall y \in \y,$
$$K(x,x)y=\sum_{j,\l} \left\langle h_j^x,h_\l^x \right\rangle_\h\left\langle y,e_j^x \right\rangle_\y e_\l^x. $$
Thus, $\forall i \in \N$
\begin{equation}\nonumber
\begin{aligned}
K(x,x)e_i^x&=\sum_{j,\l} \left\langle h_j^x,h_\l^x \right\rangle_\h\left\langle e_i,e_j^x \right\rangle_\y e_\l^x\\
&=\sum_\l \left\langle h_i^x,h_\l^x \right\rangle_\h e_\l^x.
\end{aligned}
\end{equation}
Hence 
\begin{equation}\nonumber
\begin{aligned}
\|K(x,x)\|^2_{op} &=\sup_{i\in\N} \|K(x,x)e_i^x\|_\y^2 \\
&=\sup_{i\in\N}\sum_{j,\l} \left\langle h_i^x,h_\l^x \right\rangle_\h \left\langle h_i^x,h_j^x \right\rangle_\h \left\langle e_j^x,e_\l^x \right\rangle_\y\\
&=\sup_{i\in\N}\sum_{\l} (\left\langle h_i^x,h_\l^x \right\rangle_\h)^2\\
&\le \sup_{i\in\N}\|h_i^x\|_\h^2 \sum_{\l} \|h_\l^x\|_\h^2 \le \eta^2.
\end{aligned}
\end{equation}
\epf

As a consequence of Hypothesis \ref{Hyp K bounded}, we immediately obtain the following elementary Lemma which allows us to control $\|f(x)\|_\y$ with $\|f\|_\h$. This is crucial to the proof of our main results.
\begin{lemma}\label{Lem reprod}
Let $K$ be a Hermitian positive kernel satisfying Hypothesis \ref{Hyp K bounded}. Then $\forall f \in \h$, $\|f(x)\|_\y \le \kappa \|f\|_\h$.
\end{lemma}
\bpf
\begin{equation}\nonumber
\begin{aligned}
\|f(x)\|^2_\y &= \left\langle f(x),f(x) \right\rangle_\y\\
&= \left\langle f,K(.,x)f(x) \right\rangle_\h\\
&=\left\langle K(.,x)^* f,f(x) \right\rangle_\y=\left\langle K(x,.) f,f(x) \right\rangle_\y\\
&=\left\langle K(x,x)f,f \right\rangle_\h\\
&\le \|f\|^2_\h \|K(x,x)\|_{op}\le \kappa^2 \|f\|^2_\h
\end{aligned}  
\end{equation} 
\epf

Moreover, in order to avoid measurability problems, we assume that, $\forall y_1,y_2 \in \y$, the application :
\begin{center}
\begin{math}
\begin{aligned}
\x \times \x &\longrightarrow \R \\
(x_1,x_2) &\longrightarrow \left\langle K(x_1,x_2)y_1,y_1 \right\rangle_\y,
\end{aligned}
\end{math}
\end{center}
is measurable. Since $\h$ is separable, this implies that all the functions used in this paper are measurable (for more details see \cite{VitoCapo2006}).

A regularized multi-task kernel based learning algorithm with respect to a loss function $c$ is the function defined by:
\begin{equation}
\label{AlgoMultiTask}
\begin{aligned}
\bigcup_{n \in \N} (\x \times \y)^n &\rightarrow \h\\
Z &\rightarrow f_Z,
\end{aligned}
\end{equation}
where $f_Z$ is determined by equation \eqref{definition fZ}. This leads us to introduce our second hypothesis.
\begin{hypothesis} \label{Hyp fZ well defined}
The minimization problem defined by \eqref{definition fZ} is well posed. In other words, the function $f_Z$ exists for all $Z$ and is unique.
\end{hypothesis}

Now, let us recall the notion of uniform stability of an algorithm.
\begin{definition} 
An algorithm $Z \rightarrow f_Z$ is said to be $\beta$ uniformly stable if and only if: $\forall m \ge 1 $, $\forall 1 \le i \le m,$  $\forall Z $ a training set, and $ \forall (x,y) \in \x \times \y$ a realisation of $(X,Y)$ $Z$-independent,
$$\vert c(y,f_Z,x) - c(y,f_\zi,x)\vert \le \beta.$$ 
\end{definition}
From now and for the rest of the paper, a $\beta$-stable algorithm will refer to the uniform stability. We make the following assumption regarding the loss function.
\begin{hypothesis} \label{Hyp on c}
The application $(y,f,x)\rightarrow c(y,f,x) $ is $C$-admissible, i.e.  convex with respect to $f$ and Lipschitz continuous with respect to $f(x)$, with $C$ its Lipschitz constant.
\end{hypothesis}
The above three hypotheses are sufficient to prove the $\beta$-stability for a family of multi-task kernel regression algorithms. However, to show their consistency we need an additional hypothesis.
\begin{hypothesis}\label{c control}
$\exists M>0$ such that $\forall (x,y)$ a realization of the couple $(X,Y)$, and $\forall Z$ a training set,
$$c(y,f_Z,x)\le M.$$
\end{hypothesis}
Note that the Hypotheses \ref{Hyp fZ well defined}, \ref{Hyp on c} and \ref{c control} are the same as the ones used in \cite{Bousquet}. Hypothesis \ref{Hyp K bounded} is a direct extension of the assumption on the scalar-valued kernel to multi-task setting. 


\section{Stability of Multi-Task Kernel Regression}\label{sectionStability}

In this section, we state a result concerning the uniform stability of regularized multi-task kernel regression. This result is a {direct} extension of \hbox{Theorem 22} in~\cite{Bousquet} to the case of infinite-dimensional output spaces. 
It is worth pointing out that its proof does not differ much from the scalar-valued case, and requires only small modifications of the original to fit the operator-valued kernel approach. For the convenience of the reader, we present here the proof taking into account these modifications.

\begin{theorem} \label{beta stable}
Under the assumptions \ref{Hyp K bounded}, \ref{Hyp fZ well defined} and \ref{Hyp on c}, the regularized multi-task kernel based learning algorithm $A: Z \rightarrow f_Z$ defined in~(\ref{AlgoMultiTask}) is $\beta$ stable with $$\beta= \frac{C^2 \kappa^2 }{2m\lambda}.$$
\end{theorem}
\bpf
Since $c$ is convex with respect to $f$, we have $\forall 0 \le t \le 1$
\begin{equation}\nonumber
\begin{aligned}
c(y,f_Z + t(f_\zi-f_Z),x) -c(y,f_Z,x)  \le t \left(c(y,f_\zi,x) - c(y,f_Z,x) \right).
\end{aligned}
\end{equation}
Then, by summing over all couples $(x_k,y_k)$ in $\zi$,
\begin{equation}\label{eq Rempun}
\begin{aligned}
R_{emp}(f_Z + t(f_\zi-f_Z),\zi) -R_{emp}(f_Z ,\zi)  \le t \left(R_{emp}(f_\zi,\zi) -R_{emp}(f_Z ,\zi) \right).
\end{aligned}
\end{equation}
Symmetrically, we also have
\begin{equation}\label{eq Rempdeux}
\begin{aligned}
R_{emp}(f_\zi + t(f_Z-f_\zi),\zi) -R_{emp}(f_\zi ,\zi)  \le t \left(R_{emp}(f_Z,\zi) -R_{emp}(f_\zi ,\zi) \right).
\end{aligned}
\end{equation}
Thus, by summing \eqref{eq Rempun} and \eqref{eq Rempdeux}, we obtain
\begin{equation}\label{eq Remptrois}
\begin{aligned}
&R_{emp}(f_Z + t(f_\zi-f_Z),\zi) -R_{emp}(f_Z ,\zi)\\
&\hspace{1cm} + R_{emp}(f_\zi + t(f_Z-f_\zi),\zi) -R_{emp}(f_\zi ,\zi)  \le 0.
\end{aligned}
\end{equation}
Now, by definition of $f_Z$ and $f_\zi$,
\begin{equation}\label{eq fmin}
\begin{aligned}
&R_{reg}(f_Z ,Z) - R_{reg}(f_Z + t(f_\zi-f_Z),Z)\\
&\hspace{1cm} +R_{reg}(f_\zi ,\zi)-  R_{reg}(f_\zi + t(f_Z-f_\zi),\zi)  \le 0.
\end{aligned}
\end{equation}
By using \eqref{eq Remptrois} and \eqref{eq fmin} we get
\begin{equation}\nonumber
\begin{aligned}
&c(y_i,f_Z,x_i) -c(y_i,f_Z + t(f_\zi-f_Z),x_i)\\
&\hspace{1cm} +m \lambda \left( \| f_Z \|^2_\h - \| f_Z + t(f_\zi-f_Z)\|^2_\h
+ \| f_\zi \|^2_\h -\| f_\zi + t(f_Z-f_\zi) \|^2_\h \right)\le 0,
\end{aligned}
\end{equation}
hence, since $c$ is $C$-Lipschitz continuous with respect to $f(x)$, and the inequality is true $\forall t \in \left[0,1\right]$,
\begin{equation}\nonumber
\begin{aligned}
 \|f_Z-f_\zi \|^2_\h & \le  \frac{1}{2t}  \left( \| f_Z \|^2_\h - \| f_Z + t(f_\zi-f_Z)\|^2_\h + \| f_\zi \|^2_\h -\| f_\zi + t(f_Z-f_\zi) \|^2_\h \right)\\
&\le \frac{1}{2tm\lambda} \left( c(y_i,f_Z + t(f_\zi-f_Z),x_i)-c(y_i,f_Z,x_i)\right) \\
&\le \frac{C }{2m\lambda} \|f_\zi(x_i) - f_Z(x_i)\|_\y\\
&\le  \frac{C \kappa }{2m\lambda} \|f_\zi- f_Z\|_\h,
\end{aligned}
\end{equation}
which gives that $$\displaystyle \|f_Z-f_\zi \|_\h \le  \frac{C \kappa }{2m\lambda}.$$
This implies that, $\forall (x,y)$ a realization of $(X,Y)$,
\begin{equation}\nonumber
\begin{aligned}
\vert c(y,f_Z,x) - c(y,f_\zi,x) \vert &\le C \|f_Z(x) - f_\zi(x)\|_\y\\
&\le C \kappa \|f_Z-f_\zi \|_\h \\
&\le \frac{C^2 \kappa^2 }{2m\lambda}
\end{aligned}
\end{equation}
\epf

Note that the $\beta$ obtained in Theorem \ref{beta stable} is a $O( \frac{1}{m})$. This allows to prove the consistency of the multi-task kernel based estimator from a result of \cite{Bousquet}.

\begin{theorem}\label{consistence}

Let $Z \rightarrow f_Z$ be a $\beta$-stable algorithm, whose cost function $c$ satisfies Hypothesis \ref{c control}. Then, $\forall m \ge 1$, $\forall 0 \le \delta\le 1$, the following bound holds :

\begin{equation}\nonumber
\begin{aligned}
&\P\left( \E( c(Y,f_Z,X)) \le R_{emp}(f_Z,Z)+2\beta+(4 m \beta + M)\sqrt{\frac{\ln(1/\delta)}{2m}}\right)  \ge  1- \delta . 
\end{aligned}
\end{equation}
\end{theorem}
\bpf See theorem 12 in \cite{Bousquet}.
\epf

Since the right term of the previous inequality tends to $0$ when $m \rightarrow \infty$, theorem \ref{consistence} proves the consistency of a class of multi-task kernel regression methods even when the dimensionality of the output space is infinite. 
We give in the next section several examples to illustrate the above results.


\section{Stable Multi-Task Kernel Regression Algorithms}
\label{section::StableAlgo}

In this section, we show that multi-task extension of  a number of existing kernel-based regression methods    exhibit good uniform stability properties. In particular, we focus on functional ridge \hbox{regression (RR) \cite{kadri10}}, vector-valued support vector \hbox{regression (SVR) \cite{Brudnak06}}, and multi-task logistic \hbox{regression (LR) \cite{Zhu02}}. We assume in this section that all of these algorithms satisfy Hypothesis \ref{Hyp fZ well defined}. \\[0.2cm]
\textbf{Functional response RR.} It is an extension of ridge regression (or regularized least squares regression) to functional data analysis domain \cite{kadri10}, where the goal is to predict a functional response by considering the output as a single function observation rather than a collection of individual observations. The operator-valued kernel RR algorithm is linked to the square loss function, and is defined as follows:
$$ \arg\min\limits_{f \in \h} \frac{1}{m} \sum\limits_{i=1}^m   \|y - f(x) \|_\y^ 2 + \lambda \| f \|_\h^2.$$
We should note that Hypothesis~\ref{Hyp on c} is not satisfied in the least squares context. However, we will show that the following hypothesis is a sufficient condition to prove the stability when Hypothesis \ref{Hyp K bounded} is verified (see Lemma~\ref{Hyp LSR}).
\begin{hypothesis}\label{Hyp Y bounded}
$\exists C_y>0$ such that $\|Y\|_\y<C_y$ a.s.
\end{hypothesis}
%
%
%
\begin{lemma}\label{Hyp LSR} 
Let $c(y,f,x)= \|y - f(x) \|_\y^ 2$. If Hypotheses \ref{Hyp K bounded} and \ref{Hyp Y bounded} hold, then

$$\vert c(y,f_Z,x) -  c(y,f_\zi,x)\vert \le C \|f_Z(x)-f_\zi(x)\|_\y ,$$
with $\displaystyle C=2 C_y ( 1 + \frac{\kappa}{\sqrt{\lambda}})$.
\end{lemma}
It is important to note that this Lemma can replace the Lipschitz property of $c$ in the proof of Theorem \ref{beta stable}.

{\noindent{\sc Proof of Lemma \ref{Hyp LSR}} : }
First, note that $c$ is convex with respect to its second argument. Since $\h$ is a vector space, $0 \in \h$. Thus,  
\begin{equation}\label{astuce 0}
\begin{aligned}
  \lambda \|f_Z\|^2 &\le R_{reg}(f_Z,Z) \le  R_{reg}(0,Z) \\
  &\le \frac{1}{m}\sum_{k=1}^m \|y_k \|^2\\
  &\le C_y ^2, 
\end{aligned}
\end{equation}
where the first line follows from the definition of $f_Z$ (see Equation~\ref{definition fZ}), and in the third line we used the bound on $Y$. This inequality is uniform over $Z$, and thus holds for $f_\zi$.

Moreover,
$\forall x \in \x$, 
\begin{equation}\nonumber
\begin{aligned}
\|f_Z(x)\|_\y^2 &= \left\langle f_Z(x), f_Z(x)\right\rangle_\y \\
&=\left\langle K(x,x) f_Z, f_Z\right\rangle_\h \\
&\le \|K(x,x)\|_{op} \|f_Z\|^2_\h \le \kappa^2 \frac{C_y^ 2}{\lambda} .
\end{aligned}
\end{equation}
Hence,
\begin{equation}\label{RR hyp4}
\begin{aligned}
 \|Y - f_Z(X)\|_\y &\le  \|Y \|_\y +  \|f_Z(X)\|_\y \\
&\le C_y + \kappa \frac{C_y}{\sqrt{\lambda}},
\end{aligned}
\end{equation}
where we used Lemma \ref{Lem reprod} and (\ref{astuce 0}), then
\begin{equation}\nonumber
\begin{aligned}
&\vert \|y - f_Z(x)\|_\y^2 -  \|y - f_\zi(x)\|_\y^2 \vert\\
&= \vert \|y - f_Z(x)\|_\y -  \|y - f_\zi(x)\|_\y \vert \times \vert \|y - f_Z(x)\|_\|y +  \|y - f_\zi(x)\|_\y  \vert\\
&\le 2 C_y ( 1 + \frac{\kappa}{\sqrt{\lambda}}) \|f_Z(x)-f_\zi(x)\|_\y .
\end{aligned}
\end{equation}

\epf

Hypothesis \ref{c control} is also satisfied with $M= (C/2)^2$. We can see that from Equation \ref{RR hyp4}.
Using Theorem \ref{beta stable}, we obtain that the RR algorithm is $\beta$-stable with $$\beta=\displaystyle\frac{2 C^2_y \kappa^2 ( 1 + \frac{\kappa}{\sqrt{\lambda}})^2 }{m\lambda},$$ and one can apply Theorem \ref{consistence} to obtain the following generalization bound, with probability at least $1- \delta$ : 
\begin{equation}\nonumber
\begin{aligned}
\E( c(Y,f_Z,X)) \le R_{emp}&(f_Z,Z)+\frac{1}{m}\frac{4 C^2_y \kappa^2 ( 1 + \frac{\kappa}{\sqrt{\lambda}})^2 }{\lambda}\\
&+\sqrt{\frac{1}{m}}C^2_y ( 1 + \frac{\kappa}{\sqrt{\lambda}})^2)( \frac{8  \kappa^2}{\lambda} + 1)\sqrt{\frac{\ln(1/\delta)}{2}} . 
\end{aligned}
\end{equation}
 \\[0.1cm]
\textbf{Vector-valued SVR.} It was introduced in  \cite{Brudnak06} to learn a function $f: \mathbb{R}^n \rightarrow \mathbb{R}^d$ which maps inputs $x\in  \mathbb{R}^n$ to vector-valued outputs $y\in  \mathbb{R}^d$, where $d$ is the number of tasks. In the paper, only the finite-dimensional output case was addressed, but  a general class of loss functions associated with the $p$-norm of the error  was studied. 
 In the spirit of the scalar-valued SVR, the $\epsilon$-insensitive loss function which was considered  has the following form: $ c(y,f,x)= \big | \|y - f(x) \|_p \big |_\epsilon =  \max (\|y - f(x) \|_p - \ep, 0)$, and from this general form of the $p$-norm formulation,  the special cases of 1-, 2- and $\infty$-norms was discussed.
Since in our work we mainly focus our attention to the general case of any infinite-dimensional  Hilbert space $\y$, we consider here the following  vector-valued SVR algorithm:
$$ \arg\min\limits_{f \in \h} \frac{1}{m} \sum\limits_{i=1}^m c(y_i,f,x_i)+ \lambda \| f \|_\h^2,$$
where the associated loss function is defined by:
$$ c(y,f,x) = \big | \|y - f(x) \|_\y \big |_\epsilon =  \left\{
\begin{array}{ll}
  0                            & \quad if  \|y - f(x) \|_\y \leq  \epsilon \\
 \|y - f(x) \|_\y - \ep & \quad otherwise.
\end{array}
\right.$$
This algorithm satisfies Hypothesis~\ref{Hyp on c}.  Hypothesis~\ref{c control} is also verified with $M=C_y ( 1 + \frac{\kappa}{\sqrt{\lambda}})$ when Hypothesis \ref{Hyp Y bounded} holds. This can be proved by the same way as the RR case. 
Theorem~\ref{beta stable} gives that the vector-valued SVR algorithm is $\beta$-stable with $$\beta=\displaystyle\frac{ \kappa^2 }{2m\lambda}.$$ We then obtain the following generalization bound, with probability at least $1- \delta$:
\begin{equation}\nonumber
\begin{aligned}
 \E( c(Y,f_Z,X)) \le R_{emp}(f_Z,Z)+\frac{1}{m} \frac{ \kappa^2 }{\lambda}+\sqrt{\frac{1}{m}}(\frac{2 \kappa^2 }{\lambda} + C_y ( 1 + \frac{\kappa}{\sqrt{\lambda}}))\sqrt{\frac{\ln(1/\delta)}{2}}.
\end{aligned}
\end{equation}
 \\[0.1cm]
\textbf{Multi-task LR.} As in the case of SVR, kernel logistic regression \cite{Zhu02} can be extended to the multi-task learning setting. The logistic loss can then be expanded in the manner of the  $\epsilon$-insensitive loss , that is $c(y,f,x)=\ln\left( 1 + e^{-\left\langle y,f(x) \right\rangle_\y}\right)$. It is easy to see that the multi-task LR algorithm satisfies Hypothesis~\ref{Hyp on c} with $C=1$ and Hypothesis~\ref{c control} since  $c(y,f,x) \le \ln(2)$. Thus the algorithm is $\beta$-stable with $$\beta=\displaystyle\frac{ \kappa^2 }{2m\lambda}.$$ 
The associated generalization bound, with probability at least $1- \delta$, is :
\begin{equation}\nonumber
\begin{aligned}
 \E( c(Y,f_Z,X)) \le R_{emp}(f_Z,Z)+\frac{1}{m} \frac{ \kappa^2 }{\lambda}+\sqrt{\frac{1}{m}}(\frac{2 \kappa^2 }{\lambda} + \ln(2) )\sqrt{\frac{\ln(1/\delta)}{2}}.
\end{aligned}
\end{equation}

Hence, we have obtained generalization bounds for the RR, SVR and LR algorithms even when the kernel does not satisfy the Hilbert Schmidt \hbox{property (see} the following section for examples of such kernels).


\section{Discussion and Examples}\label{sectionExample}

We provide generalization bounds for multi-task kernel regression when the output space is infinite dimensional Hilbert space using the notion of algorithmic stability. 
 As far as we are aware, the only previous study of this problem was carried out in \cite{VitoCapo2006}. However, only learning rates of the regularized least squares algorithm was provided when the operator-valued kernel is assumed to be Hilbert-Schmidt.  
We have shown in Section~\ref{sectionStability} that one may use non Hilbert-Schmidt kernels in addition to obtaining theoretical guarantees. It should be pointed out that in the finite-dimensional case the Hilbert-Schmit assumption is always satisfied, so it is important to discuss applied machine learning situations where infinite-dimensional output spaces can be encountered. Note that our bound can be recovered from \cite{VitoCapo2006} when both our and their hypotheses are satisfied  
\\[0.3cm]
\textbf{Functional regression.} From a functional data analysis (FDA) point of view, infinite-dimensional output spaces for operator estimation problems are frequently encountered in functional response regression analysis, where the goal is to predict an entire function. FDA is an extension of multivariate data analysis suitable when the data are curves, see \cite{ramsay05} for more details.  A functional response regression problem takes the form $y_i = f(x_i)+ \epsilon_i$ where both predictors $x_i$ and responses $y_i$ are functions in some functional Hilbert space, most often the space $L^2$ of square integrable functions. In this context, the function $f$ is an operator between two infinite-dimensional Hilbert spaces. Most previous work on this model suppose that the relation between functional responses and predictors is linear. The functional regression model is an extension of the multivariate linear model and has the following form: $$ y(t) = \alpha(t) + \beta(t) x(t) + \epsilon(t) $$ for a regression parameter $\beta$. In this setting, an extension to nonlinear contexts can be found in \cite{kadri10} where the authors showed how Hilbert spaces of function-valued functions and infinite-dimensional operator-valued reproducing kernels can be used as a theoretical framework to develop nonlinear functional regression methods. A multiplication based operator-valued kernel was proposed, since the linear functional regression model is based on the multiplication operator.
\\[0.3cm]
\textbf{Structured output prediction.} One approach to dealing with this problem is kernel dependency estimation (KDE) \cite{Weston03}. It is based on defining a scalar-valued kernel $k_\y$ on the outputs, such that one can transform the problem of learning a mapping between input  data $x_i$ and structured outputs $y_i$ to a problem of learning a Hilbert space valued function $f$ between $x_i$ and $\Phi(y_i)$, where $\Phi(y_i)$ is the projection of $y_i$ by $k_\y$ into a real-valued RKHS $\mathcal{H}_\y$. Depending on the kernel $k_\y$, the RKHS $\mathcal{H}_\y$ can be infinite-dimensional. In this context, extending KDE for RKHS with multi-task kernels was first introduced in \cite{brouard11}, where an identity based operator-valued kernel was used to learn the function $f$.
\\[0.3cm]
\textbf{Conditional mean embedding.} 
As in the case of structured output learning, the output space in the context of conditional mean embedding is a scalar-valued RKHS.  In the framework of probability distributions embedding, Gr\"unew\"alder et al. \cite{arthur12} have shown an equivalence between RKHS embeddings of conditional distributions and multi-task kernel regression.  On the basis of this link, the authors derived a sparse embedding algorithm using the identity based operator-valued kernel.
\\[0.3cm]
\textbf{Collaborative filtering.} The goal of collaborative filtering (CF) is to build a model to predict preferences of clients \textit{“users”} over a range of products \textit{“items”} based on information from customer’s past purchases. In \cite{Abernethy09}, the authors show that several CF methods such as rank-constrained optimization, trace-norm regularization, and those based on Frobenius norm regularization, can all be cast as special cases of spectral regularization on operator spaces. Using operator estimation and spectral regularization as a framework for CF permit to use potentially more information and incorporate additional \textit{user-item} attributes to predict preferences. A generalized CF approach consists in
learning a preference function $f(\cdot,\cdot)$ that takes the form of a linear operator from a Hilbert space of users to a Hilbert space of items, $f(\cdot,\cdot) = \langle x , F y \rangle$ for some compact operator $F$.
\\[0.3cm]
Now we want to emphasize that in the case of infinite dimensions Hypothesis~\ref{Hyp K bounded} on the kernel is not equivalent to that used in \cite{VitoCapo2006}. We have shown in Section~\ref{sectionHypothesis} that our assumption on the kernel is weaker. To illustrate this, we provide below examples of operator-valued kernels which satisfy Hypothesis \ref{Hyp K bounded} but are not Hilbert-Schmidt, as was assumed in \cite{VitoCapo2006}.

\begin{example}
Identity operator. Let $c>0$, $d\in \N^ *$, $\x = \R^d$, $I$ be the identity morphism in $L(\y)$, and $K(x,t)=\exp( -c\|x-t\|_2^ 2) \times I$.The kernel $K$ is positive, Hermitian, and   

\begin{equation}\label{exemple ID}
\begin{aligned}
\| K(x,x) \|_{op}= \|I \|_{op}= 1\\
 Tr(K(x,x))= Tr(I)= + \infty.
 \end{aligned}
\end{equation}
This result is also true for any kernel which can be written as  $K(x,t)=k(x,t) I$, where $k$ is a positive-definite scalar-valued kernel satisfying that $\sup_{x \in \x} k(x,x) < \infty$. Identity based operator-valued kernels are already used in structured output learning~\cite{brouard11} and conditional mean embedding~\cite{arthur12}.
\end{example}

\begin{example}
Multiplication operator - Separable case. Let $k$ be a positive-definite scalar-valued such that $\sup_{x \in \x} k(x,x) < \infty$, $\I$ an interval of $\R$, $C>0$, and $\y~=~L^2(\I,\R)$. Let $ f \in L^ \infty (\I,\R)$ be such that $\|f\|_\infty< C$.

We now define the multiplication based operator-valued kernel $K$  as follows 
$$K(x,z)y(.)= k(x,z) f^2(.) y(.) \in \y.$$ 
Such kernels are suited to extend linear functional regression to nonlinear context~\cite{kadri10}. 
$K$ is a positive Hermitian kernel but, even if $K$ always satisfy Hypothesis \ref{Hyp K bounded}, the Hilbert-Schmidt property of $K$ depends on the choice of $f$ and may be difficult to verify. For instance, let $f(t)= \frac{C}{2} (exp ( -t^2) +1)$. Then

\begin{equation}\label{exemple multi}
\begin{aligned}
\| K(x,x) \|_{op} &\le C^2 k(x,x) \\
 Tr(K(x,x))&= \sum_{i \in \N} \left\langle K(x,x) y_i,y_i \right\rangle  \ge k(x,x) \frac{C}{2} \sum_{i \in \N} \|y_i\|_2^2 = \infty,
 \end{aligned}
\end{equation}
where $(y_i)_{i \in \N}$ is an orthonormal basis of $Y$ (which exists, since $Y$ is separable).

\end{example}

\begin{example}Multiplication Operator - Non separable case\footnote{A kernel $K(x,t)$ is called non separable, as opposed to separable, if it cannot be written as the product of a scalar valued kernel $k(x,t)$ and a $L(\y)$ operator independent of the choice of $x,t\in\x$.}. Let $\I$ an interval of $\R$, $C~>~0$, $\y~=~L^2(\I,\R)$, $\x~=\left\{ f\in L^\infty(\I,\R)\text{ such that }\|f\|_\infty \le C \right\}$.

Let $K$  be the following operator-valued function:
$$K(x,z)y(.)= x(.)z(.) y(.) \in \y.$$ $K$ is a positive Hermitian kernel satisfying Hypothesis \ref{Hyp K bounded}. Indeed,
\begin{equation}\nonumber
\begin{aligned}
\| K(x,x) \|_{op} = \max_{y \in \y} \frac{\sqrt{\int_\I x^4(t) y^2(t)dt} }{\|y\|_2} \le C^2 \max_{y \in \y} \frac{\sqrt{\int_\I y^2(t)dt }}{\|y\|_2} \le C^2.
 \end{aligned}
\end{equation}
On the other hand, $K(x,x)$ is not Hilbert-Schmidt for all choice of $x$ (in fact it is not Hilbert Schmidt as long as $\exists \ep>0$ such that $\forall t\in \I$, $x(t)> \ep >0$). To illustrate this, let choose $x=\frac{C}{2} (exp ( -t^2) +1)$ as defined in the previous example. Then, for $(y_i)_{i \in \N}$ an orthonormal basis of $Y$, we have 
\begin{equation}\nonumber
\begin{aligned}
 Tr(K(x,x))&= \sum_{i \in \N} \left\langle K(x,x) y_i,y_i \right\rangle  = \sum_{i \in \N} \int_\I x^2(t) y_i^2(t)dt \\
 & \ge \frac{C}{2} \sum_{i \in \N} \int_\I y_i^2(t)dt =  \frac{C}{2} \sum_{i \in \N} \|y_i\|_2^2 = + \infty .
 \end{aligned}
\end{equation}

\end{example}

\begin{example} Sum of kernels. This example is provided to show that in the case of multiple kernels the sum of a non Hilbert-Schmidt kernel with an Hilbert-Schmidt one gives a non-Hilbert-Schmidt kernel. This makes the assumption on the kernel of \cite{VitoCapo2006} inconvenient for multiple kernel learning (MKL) \cite{kadri12}, since one would like to learn a combination of different kernels which can be non Hilbert-Schmidt (like the basic identity based operator-valued kernel).

Let $k$ be a positive-definite scalar-valued kernel satisfying $\sup_{x \in \x} k(x,x) < \infty$, $\y$ a Hilbert space, and $y_0 \in \y$. 
Let $K$ be the following kernel: 
$$K(x,z)y= k(x,z) \left( y + \left\langle y,y_0 \right \rangle y_0 \right). $$
$K$ is a positive and Hermitian kernel. Note that a similar kernel was proposed for multi-task learning \cite{kadri12}, where the identity operator is used to encode the relation between a task and itself, and a second kernel is added for sharing the information between tasks. $K$ satisfies Hypothesis \ref{Hyp K bounded}, since
\begin{equation}\nonumber
\begin{aligned}
\| K(x,x) \|_{op} = \max_{y \in \y, \|y\|_\y=1} \|k(x,x) \left( y + \left\langle y,y_0 \right \rangle y_0 \right) \|_\y \le \vert k(x,x) \vert (1 + \|y_0\|_\y^2).
 \end{aligned}
\end{equation}
However, $K$ is not Hilbert Schmidt. Indeed, it is the sum of a Hilbert Schmidt kernel ( resp. $K_1(x,z)y=k(x,z) \left\langle y,y_0 \right \rangle y_0$) and a Hilbert-Schmidt one ( resp. $K_2(x,z)y=k(x,z) y$), which is not Hilbert Schmidt. To see this, note that since the trace of $K_1$ is the sum over an absolutely summable family, and the trace is linear, so the trace of $K$ is the sum of an convergent series and a divergent one, hence it diverges, so $K$ is not Hilbert Schmidt.
\end{example}

\section{Conclusion}\label{sectionConclusion}

We have shown that a large family of multi-task kernel regression algorithms, including functional ridge regression and vector-valued SVR, are $\beta$-stable even when the output space is infinite-dimensional. This result allows us to provide generalization bounds and to prove under mild assumptions on the kernel the consistency of these algorithms.
However, obtaining learning bounds with optimal rates  for infinite-dimensional multi-task kernel based algorithms is still an open question.

\bibliographystyle{splncs}
\bibliography{biblioinfo}

\end{document}